\documentclass{article}

\usepackage{arxiv}
\usepackage{float} 
\usepackage[table]{xcolor}  
\usepackage[utf8]{inputenc} 
\usepackage[T1]{fontenc}    
\usepackage{hyperref}       
\usepackage{url}            
\usepackage{booktabs}       
\usepackage{amsfonts}       
\usepackage{nicefrac}       
\usepackage{microtype}      
\usepackage{lipsum}
\usepackage{graphicx}
\graphicspath{ {./images/} }

\title{Monkey Transfer Learning Can Improve Human Pose Estimation}

\author{
Bradley Scott \\
School of Medicine, Medical Sciences and Nutrition \\
University of Aberdeen \\
Aberdeen, United Kingdom \\
\texttt{b.scott.20@abdn.ac.uk} \\
\And
Clarisse de Vries \\
School of Health and Wellbeing \\
University of Glasgow \\
Glasgow, United Kingdom \\
\texttt{Clarisse.deVries@glasgow.ac.uk} \\
\And
Aiden Durrant \\
School of Natural and Computing Sciences \\
University of Aberdeen \\
Aberdeen, United Kingdom \\
\texttt{aiden.durrant@abdn.ac.uk} \\
\And
Nir Oren \\
School of Natural and Computing Sciences \\
University of Aberdeen \\
Aberdeen, United Kingdom \\
\texttt{n.oren@abdn.ac.uk} \\
\And
Edward Chadwick \\
School of Engineering \\
University of Aberdeen \\
Aberdeen, United Kingdom \\
\texttt{edward.chadwick@abdn.ac.uk} \\
\And
Dimitra Blana \\
School of Medicine, Medical Sciences and Nutrition \\
University of Aberdeen \\
Aberdeen, United Kingdom \\
\texttt{dimitra.blana@abdn.ac.uk} \\
}
\begin{document}
\maketitle
\begin{abstract}
In this study, we investigated whether transfer learning from macaque monkeys could improve human pose estimation. Current state-of-the-art pose estimation techniques, often employing deep neural networks, can match human annotation in non-clinical datasets. However, they underperform in novel situations, limiting their generalisability to clinical populations with pathological movement patterns. Clinical datasets are not widely available for AI training due to ethical challenges and a lack of data collection. We observe that data from other species may be able to bridge this gap by exposing the network to a broader range of motion cues. We found that utilising data from other species and undertaking transfer learning improved human pose estimation in terms of precision and recall compared to the benchmark, which was trained on humans only. Compared to the benchmark, fewer human training examples were needed for the transfer learning approach (1,000 vs 19,185). These results suggest that macaque pose estimation can improve human pose estimation in clinical situations. Future work should further explore the utility of pose estimation trained with monkey data in clinical populations.
\end{abstract}


\section{Introduction}

Pose estimation has applications across multiple domains such as entertainment, sports performance, clinical gait studies, and rehabilitation \cite{zheng_deep_2023}. The deep learning based methods used in pose estimation algorithms can achieve performance that matches human annotation; however, training them requires large annotated datasets consisting of tens of thousands of images \cite{desmarais_review_2021, ben_gamra_review_2021}. These datasets can be collated from publicly available video sources, such as YouTube in the case of MPII \cite{andriluka_2d_2014}, relying on hand annotated images. Alternatively, data can be captured in laboratory settings, as with Human 3.6M \cite{han_space-time_2017}, which provide more precise ground truth annotations but limited participant and environmental diversity. Despite matching human annotation, state-of-the-art methods underperform in novel situations, particularly in applications involving  clinical populations with pathological movement patterns \cite{pardell_movement_2024, galna_accuracy_2014, horsak_concurrent_2023}. Existing datasets rarely include individuals with such conditions, often due to a lack of data collection \cite{philp_international_2022} and ethical challenges associated with making clinical data publicly available. This omission results in pose estimation algorithms that perform worse on those with movement disorders compared to healthy controls.

One of the main challenges in machine learning is the scarcity of labelled data for a new task. Transfer learning \cite{yang_transfer_2020} addresses this issue by taking a model developed for one problem and reusing it as a starting point for another, adapting its learned knowledge to the target scenario. This approach is particularly useful in deep learning, where large, annotated datasets are often needed to achieve high performance. For example, ImageNet \cite{deng_imagenet_2009} is used as a source task to pre-train models on a large set of labelled images, and then these pre-trained models are fine-tuned to improve feature identification for various target tasks such as object detection and segmentation \cite{huh_what_2016}. Transfer learning has been successfully used to improve clinical pose estimation by initially training AI with typical gait, and subsequently applying additional training data from various disease populations, such as children with bone disease \cite{vafadar_novel_2021}. This supports the hypothesis that a lack of data inclusion is a significant detriment to accuracy in clinical pose estimation and that feature identification is less of an issue than movement characteristics being represented in the dataset. Given the wide range of pathologies and barriers for widely accessible clinical training data, using human clinical populations for AI training is unlikely to be commonplace in the near future. Instead, we can look to other species to provide the diversity of poses required to bridge this gap.

Recently, DeepLabCut has implemented transfer learning across various animals, including dogs and mice, to enhance pose estimation accuracy and develop species agnostic models \cite{ye_superanimal_2024}. One example is their quadrupedal model, designed to make predictions for both deer and zebras. This is made possible by treating pose estimation as a panoptic segmentation task \cite{kirillov_panoptic_2019}, where a class is assigned to every pixel in the image. The model then infers keypoints from the input image by training on a collection of various heterogeneous datasets — for example, datasets where different keypoints are annotated due to species differences or differing annotation conventions. This method results in a network that learns features across species and can estimate keypoints on an image by image basis. The central idea is that shared learning between species proves useful for the task of pose estimation. 

Herein, we employ transfer learning by starting with a pose estimation network trained on a large publicly available macaque monkey dataset \cite{labuguen_macaquepose_2021}, and then further training it with human data to improve human pose estimation. This macaque network is trained on an assortment of hand-annotated macaque monkey pose images that may be leveraged to improve the accuracy of human pose estimation. Although visually different from humans, the task of monkey pose estimation involves predicting a similar set of keypoints, given the similarities in our skeletal structure. Furthermore, monkeys exhibit much more diverse movements than humans, driven by their need to climb, swing, and leap across trees.  The data collection process for monkeys is also unrestricted, allowing the collection of a wider range of movements, such as bathing. In addition, animals do not have the same restrictions around data sharing and are captured in a natural manner. We hypothesize that the variety of poses detectable by the macaque network will benefit the learning of human pose estimation. The diversity in pre-training exposes the network to various movement contexts and relationships and allows the network to form associations, making it more adaptable and effective in recognizing typical human movements even with fewer training examples.  Our research addresses the question: Can transfer learning from macaque monkeys improve human pose estimation?  

\section{Methods}

First, the macaque monkey network is used as a baseline to establish the initial performance before any enhancements. Next, the human network is set up to serve as a benchmark for evaluating the transfer learning approach. Finally, the transfer learning model is created by fine-tuning the macaque monkey network using human examples.

\subsection{Macaque Monkey Network Baseline}

The Macaque Monkey network that is used for the foundations of the transfer learning and that will ultimately act as evidence for learning, was trained on 14,697 images of single monkeys using DeepLabCut (DLC). An old version of DLC was used and as a result there are certain limitations, e.g. only single animal frames can be processed. As a workaround for multi-animal images, the monkey dataset was segmented per monkey and the segmentation mask for each monkey was included in the annotations. The segmented images were resized to adjust the length to 640 pixels while maintaining the images' aspect ratio, before inputting them into the network.

The results from the original paper \cite{labuguen_macaquepose_2021} were replicated to ensure understanding of the methods for a fair comparison. Predicted and ground truth keypoints were plotted on the monkey images to check validity. 

Then, 1000 randomly selected single person images from the MPII dataset were resized to a width of 640 pixels while maintaining aspect ratio. This resizing, as described in the original study, was accompanied by adjustments to the ground truth annotations and scaling factors to maintain consistency with the transformed image dimensions. In line with the original methodology, predictions with a confidence >= 0.4 were considered valid. Since different images have varying scales (e.g., the monkey is closer in some images and farther in others), the Euclidean distance (the difference between predicted and ground truth keypoints) in the Original Macaque paper is normalised by the length of the bounding box of the monkey. This normalisation adjusts for variations in the size of the monkey across images, meaning that the error is expressed as a fraction of the monkey's bounding box size rather than raw pixel distances. However, MPII includes a scaling factor that normalises all subjects to a 200px height, so this was utilised instead of creating segmentation masks.

\subsection{Human Network Benchmarking}

The human network used for the benchmarking comparison is the \texttt{DLC human fullbody resnet 101} model, which is a direct implementation of DeeperCut \cite{insafutdinov_deepercut_2016}. This network was chosen as a benchmark due to the similar \texttt{resnet 50} architecture used to train the macacque network. The purpose of this methodology is to demonstrate learning, not beat the state-of-the-art.

The dataset used for this comparison was MPII which was filtered to only include single person images. The version of DLC the Macaque network was trained on only supported single-animal pose estimation. Therefore, single person was used here to provide a  fair comparison. MPII provides images with various dimensions and accompanying annotations that include ground truth keypoint coordinates and a scaling factor. This scaling factor (Scaling A) is calculated using the original image dimensions and acts to normalise each person to a height of 200px, negating evaluation issues related to different image dimensions and subject framing i.e., a person being very close or far away in an image.

Images were resized to match the expected input of the network (1280 * 720 pixels) while maintaining the aspect ratio. The scaling factor (Scaling B) was stored, which scales the predicted keypoints back into the coordinate system of the original image and annotations for plotting and evaluation.

\subsection{Transfer Learning Model}

The Macaque network was trained on 1000 randomly selected single person images with a 95 percent training validation split. 

Training parameters were set to their default DLC values, with the exception of \texttt{keepdeconvweights}, which was set to false, and the output keypoints, which were configured to align with the MPII dataset. Setting \texttt{keepdeconvweights} to false allows the network to retrain over the deconvolutional layers and predict new outputs as the monkey keypoints do not match the MPII keypoints exactly. For example, there is no headtop in the macaque dataset, which is required for the normalisation of MPII. The batch size was set to 8 to accommodate the small dataset size but high complexity and extensive processing required for the images.

The network was trained for 164,000 iterations, with training halted when the loss began to plateau to prevent overfitting. While a decreasing loss is a good indicator of the model learning effectively from the training data and aligning well with the validation set, it is crucial to evaluate the model on more generalisable metrics such as RMSE (Root Mean Square Error) with a test set (the same 1000 images that were used for the benchmark). 

\subsection{Environment}
The computations were performed on the Aberdeen University High-Performance Computing system (HPC), utilising a Tesla V100-PCIE 32GB GPU node. Access to the system was established remotely via SSH, within a tmux multiplexer environment.

DeepLabCut version 2.2.0.6 was used as this was the latest version that was supported by the graphics card driver on the HPC.

\subsection{Statistical Analysis}

ROC-AUC (Receiver Operating Characteristic - Area Under the Curve) was calculated as a measure of the overall discriminative ability of the models in keypoint detection, whereas Precision, Recall, and F-score were used as measures of keypoint localization accuracy at specific thresholds. The classification thresholds for keypoint detection were evaluated using the Percentage of Correct Keypoints normalised by head segment length (PCKh@0.5). This metric determines a keypoint to be accurately predicted if it is within a threshold of 0.5 times the head segment length from its true position. 

RMSE (Root Mean Square Error) was computed for each axis (X and Y) to quantify the discrepancies between the predicted positions and the ground truth. Additionally, the Euclidean RMSE, representing the direct distance between predicted and true keypoints, was calculated to provide a comprehensive measure of overall prediction accuracy. The RMSE at the threshold corresponding to the maximum F-score was used for model comparisons to ensure that evaluations were based on each model's optimal performance point. To ensure an accurate representation of error, keypoints that were not present in the image, as indicated by an absence in the ground truth MPII annotations, were excluded from the RMSE evaluation. 

\section{Results}

\begin{table}[H]
\centering
\caption{Overall Performance Metrics for Baseline (B), Benchmark (BM), and Transfer Learning (TL) Models. In this table, cells highlighted in green denote superior performance across the compared models.}
\label{tab:tab1} 
\begin{tabular}{|c|c|c|c|}
\hline
\textbf{Metric} & \textbf{Baseline (B)} & \textbf{Benchmark (BM)} & \textbf{Transfer Learning (TL)} \\ \hline
\textbf{AUC} & 0.63 & \cellcolor{green!25}0.84 & 0.77 \\ \hline
\textbf{F1 Score} & 0.33 & 0.75 & \cellcolor{green!25}0.82 \\ \hline
\textbf{Precision} & 0.21 & 0.69 & \cellcolor{green!25}0.72 \\ \hline
\textbf{Recall} & 0.74 & 0.83 & \cellcolor{green!25}0.94 \\ \hline
\textbf{Average RMSE (px)} & 83.44 & 40.25 & \cellcolor{green!25}27.83 \\ \hline
\end{tabular}
\end{table}

Overall, the effectiveness of learning can be demonstrated by the improved results across all metrics when comparing the macaque network baseline and transfer learning model (see Table \ref{tab:tab1}). However, the utility of the model is best demonstrated when comparing the transfer learning with the benchmark model. 

\subsection{AUC Performance: Benchmark vs. Transfer Learning}

\begin{figure}[H] 
    \centering
    \includegraphics[width=0.8\textwidth]{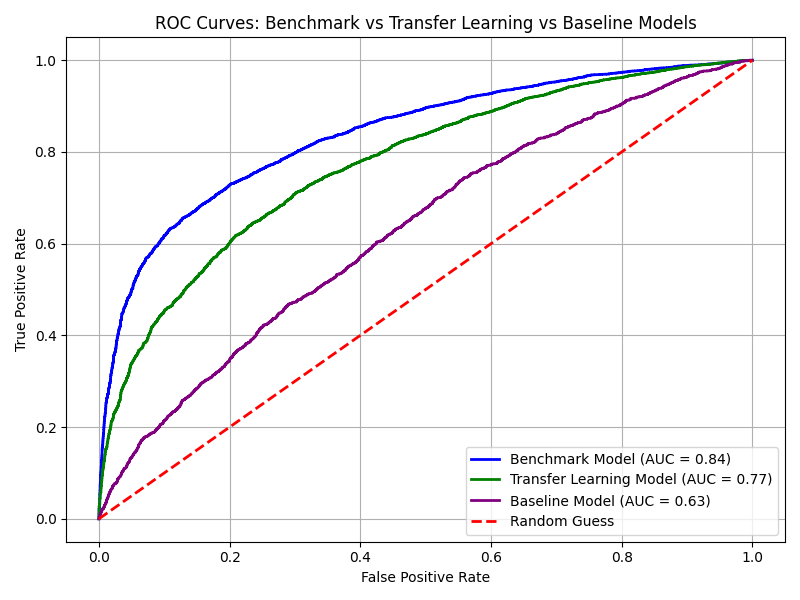} 
    \caption{ROC curves: Benchmark vs Transfer Learning Models.}
    \label{fig:fig1}
\end{figure}

ROC-AUC is a comprehensive measure of a model’s ability to distinguish between correct and incorrect predictions across various confidence thresholds (See Figure \ref{fig:fig1}). The Benchmark model achieves a higher AUC (0.84) compared to the Transfer Learning model (0.77), indicating that the Benchmark model is better at separating correct predictions (those matching the ground truth) from incorrect ones across all thresholds. This performance is particularly important when both types of errors, misclassified keypoints and missed keypoints, are equally undesirable.

\begin{figure}[htbp] 
    \centering
    \includegraphics[width=1\textwidth]{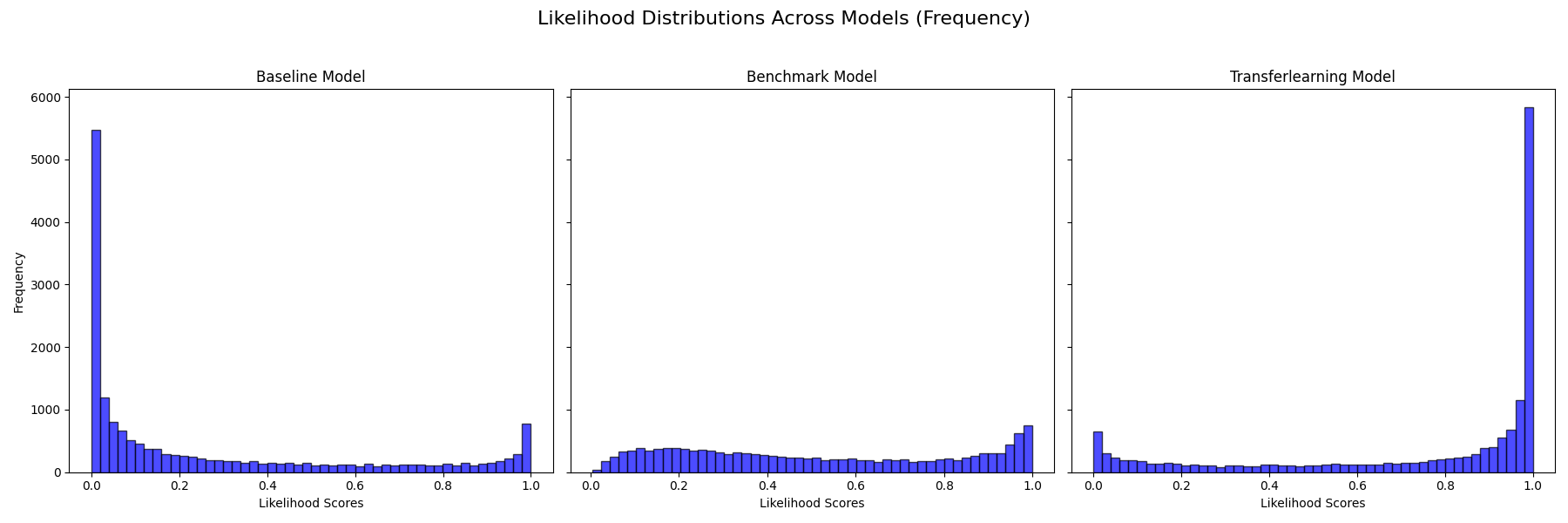} 
    \caption{Likelihood distributions: Benchmark vs Transfer Learning Models.}
    \label{fig:fig2}
\end{figure}

The Benchmark model’s higher AUC can be attributed to its broader and more balanced distribution of confidence scores (see Figure \ref{fig:fig2}), which are returned with each prediction as a value between 0 and 1. These confidence scores are more evenly distributed for the Benchmark model, allowing it to effectively distinguish correct predictions from incorrect ones at a wide range of thresholds. This consistent separation across thresholds enhances the model’s overall discriminative ability, contributing to its superior AUC.

In contrast, the Transfer Learning (TL) model achieves a higher number of correct predictions overall, with 7,207 compared to the Benchmark model’s 5,146. This reflects the TL model’s strong emphasis on capturing true keypoints, which directly contributes to its high Recall. However, the TL model’s confidence scores are more concentrated toward higher values (closer to 1), reducing the variability in scores between correct and incorrect predictions. This concentration limits its ability to separate the two classes effectively across a wide range of thresholds, leading to a lower AUC. While the TL model excels at identifying keypoints at specific thresholds, this skewed distribution impacts its overall ranking ability across thresholds.

\subsection{Superior Precision, Recall, and F1 Score in Transfer Learning}

Despite the lower AUC, the Transfer Learning model shows superior Precision, Recall, and F1 Score. These metrics are critical in evaluating the performance at specific operational points, rather than across all possible thresholds like AUC.

\textbf{F1 Score:} Reflecting the balance between Precision and Recall, the F1 Score is highest for the TL model (0.82), demonstrating its robustness in handling the trade-off between missing keypoints and incorrectly predicting new ones.

\textbf{Precision:} The TL model has a higher Precision (0.72) compared to the BM model (0.69). This suggests that the TL model's prediction of a keypoint's presence is more likely to be correct than the BM model's prediction. 

\textbf{Recall:} The TL model also exhibits a significantly higher Recall (0.94) than the BM model (0.82). This high recall reflects the model's strong ability to identify true keypoints, which is crucial in scenarios where missing a keypoint could lead to incorrect movement analysis due to the reliance on the kinematic chain.

\subsection{Practical Implications for Keypoint Estimation}

For practical applications in keypoint estimation, the choice between models depends on the specific requirements of the task rather than the theoretical superiority of one metric over another. In scenarios where missing a keypoint could lead to significant downstream errors, the superior Precision and Recall of the TL model make it more suitable, despite its lower AUC. While the Benchmark model may excel at generally distinguishing between classes, the TL model's ability to accurately identify and confirm keypoints makes it potentially more valuable for specific keypoint detection tasks, such as clinical movement analysis, where precise and reliable keypoint detection is paramount.

\section{Discussion}

Our results demonstrate that monkey pose estimation can be beneficial to human pose estimation by achieving higher precision and recall in keypoint detection compared to the benchmark model. The transfer learning approach, specifically, excels in accurately identifying more keypoints with a superior F1 score. In addition, we have demonstrated that significantly less human examples were required to train the network to perform pose estimation than the benchmark network (1,000 vs 19,185), which further demonstrates the effectiveness of learning from the macaque model. 

Exactly how convolutional neural networks (CNNs) are able to learn from monkeys to improve human pose estimation requires further research; as is the case with understanding how CNNs learn to solve computer vision tasks more generally. However, a hypothesis for this could be related to how CNNs group features and outputs. Gestalt theories are principles of perception that describe how humans naturally organize visual information into structured wholes. CNNs share some parallels with these theories (closure being commonly demonstrated) \cite{zhang_investigating_2024, kim_neural_2021, amanatiadis_understanding_2018}, as they process and structure visual data to make sense of complex information. One relevant Gestalt principle is common fate, which suggests that elements moving together (in our case with similar direction, joint mechanics, and relative positioning of body parts) are perceived as a group. The group here can be thought of as the keypoint or body part. It is likely that this concept is reflected in how CNNs learn movement patterns in data.

Even without observing movement sequentially between frames, the network can develop associations of motion patterns that imply movement and suggest likely relationships between body parts in common poses. For example, it may learn that in a running pose, an extended leg often pairs with an opposite arm swing, allowing it to expect this alignment in similar configurations between species. These associations help the CNN generalise across different poses, guiding it in accurately predicting keypoints by recognizing familiar spatial patterns even from a single frame. For example, a CNN trained only on typical human movement may struggle to correctly predict keypoints in pathological movements. As it has not encountered body parts in that context, it may struggle with different relationships between points and associated body parts, e.g. it may associate straight arms with straight legs not crouched gait. However, by applying transfer learning with a different dataset, such as one with data from monkey movement, which includes diverse joint mechanics and positioning — the network could generalise better to unfamiliar human movement patterns. Sequential methods such as vision transformers may require temporal continuity between species i.e. monkeys move much faster than humans. However, the single frame method is perhaps an advantage here as temporal continuity between movements isn't required for the pose estimation to find a solution; even the latest DLC cross-species animal models are single frame.

Overall, macaque pose estimation can be leveraged to benefit human pose estimation by providing a wider range of poses and associations of keypoints which can result in fewer human training examples required for training. However, the main limitation of this work is the restriction of using DLC for both the macaque and human networks. The success of the transfer learning is bound to the accuracy of the macaque network and also the sophistication of methods employed in transfer learning that are missing from this version of DLC, such as freezing layers and feature visualization techniques. Future work should aim to take this premise outside of DLC to better understand how monkey data can improve pose estimation in humans with movement pathologies. An exploration of how this method of transfer learning could be applied to novel clinical populations would also be beneficial to understand how it can improve the accuracy of pose estimation without specific training examples.

\bibliographystyle{unsrt}  
\bibliography{references}  






 \end{document}